%% file: paper.tex
\pdfoutput=1

\documentclass[11pt]{article}

\usepackage{acl}

\usepackage{times}
\usepackage{latexsym}

\usepackage[T1]{fontenc}

\usepackage[utf8]{inputenc}

\usepackage{microtype}

\usepackage{inconsolata}

\usepackage{graphicx}
\usepackage{multicol}
\usepackage{multirow}
\usepackage{enumitem}
\usepackage{hyperref}
\usepackage{booktabs}
\usepackage{amssymb}
\usepackage{amsmath,mathtools,xspace,color,soul}
\usepackage{subcaption}
\usepackage{appendix}
\usepackage{algorithm}
\usepackage{algpseudocode}
\usepackage[breakable,skins]{tcolorbox}
\usepackage{alltt}
\usepackage{xcolor}
\usepackage{marvosym}
\newtcolorbox{AIBox}[2][]{aibox,title=#2,#1}

\tcbset{
  aibox/.style={
    width=0.95\textwidth,
    top=5pt,
    colback=black!05,
    colframe=black!20,
    colbacktitle=black!50,
    enhanced,
    center,
    attach boxed title to top left={yshift=-0.1in,xshift=0.1in},
    boxed title style={boxrule=0pt,colframe=white,},
  }
}
\DeclareMathOperator*{\argmax}{argmax}

\title{Planning with Diffusion Models for Target-Oriented Dialogue Systems}

\author{Hanwen Du\textsuperscript{$\clubsuit$} \  Bo Peng\textsuperscript{$\clubsuit$} \  Xia Ning\textsuperscript{$\clubsuit$}\textsuperscript{$\spadesuit$}\textsuperscript{$\heartsuit$}\Letter \\
\textsuperscript{$\clubsuit$}Department of Computer Science
and Engineering, The Ohio State University, USA\\
\textsuperscript{$\spadesuit$}Department of Biomedical Informatics, The Ohio State University, USA\\
\textsuperscript{$\heartsuit$}Translational Data Analytics Institute, The Ohio State University, USA\\
\texttt{\{du.1128,peng.707,ning.104\}@osu.edu}}

\input{define}
\begin{document}
\maketitle
\begin{abstract}
Target-Oriented Dialogue (\TOD) remains a significant challenge in the LLM era, where strategic dialogue planning is crucial for directing conversations toward specific targets. However, existing dialogue planning methods generate dialogue plans in a step-by-step sequential manner, and may suffer from compounding errors and myopic actions. To address these limitations, we introduce a novel dialogue planning framework, \method, which leverages diffusion models to enable non-sequential dialogue planning. \method formulates dialogue planning as a trajectory generation problem with conditional guidance, and leverages a diffusion language model to estimate the likelihood of the dialogue trajectory. To optimize the dialogue action strategies, \method introduces three tailored guidance mechanisms for different target types, offering flexible guidance toward diverse \TOD targets at test time. Extensive experiments across three diverse \TOD settings show that \method can effectively perform non-myopic lookahead exploration and optimize action strategies over a long horizon through non-sequential dialogue planning, and demonstrates strong flexibility across complex and diverse dialogue scenarios. Our code and data are accessible through \url{https://github.com/ninglab/DiffTOD}.
\end{abstract}
\section{Introduction}
\begin{figure}
    \centering
    \includegraphics[width=\linewidth]{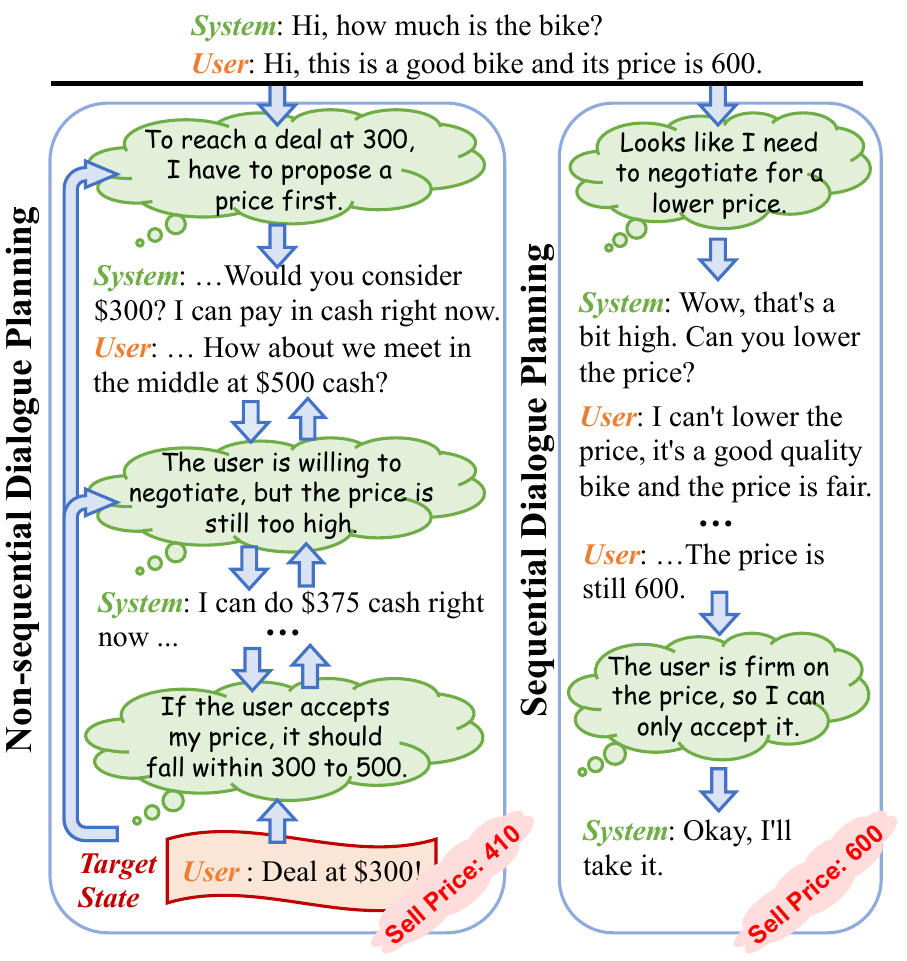}
    \caption{Using sequential and non-sequential dialogue planning methods for negotiation dialogues.}
    \label{fig:dialogue_planning}
\end{figure}
Target-Oriented Dialogue (\TOD) systems can assist users in accomplishing specific targets through interactive natural language conversations~\cite{deng2023survey,qin-etal-2023-end}, such as completing a transaction~\cite{he-etal-2018-decoupling} and providing personalized recommendations~\cite{wang-etal-2023-target}. 
With the rise of Large Language Models (LLMs), \TOD systems have undergone a paradigm shift toward LLM-integrated architectures, which are highly capable of generating high-quality, human-like responses that enhance user engagement experiences.~\cite{ou-etal-2024-dialogbench,deng2024plugandplay}.
However, since LLMs are typically trained to follow instructions passively~\cite{ouyang2022training}, they often lack the proactivity to plan and guide the conversation toward the intended target~\cite{hao-etal-2023-reasoning,deng2025survey}, a crucial property for the successful achievement of targets in \TOD~\cite{wang-etal-2023-dialogue,he-etal-2024-planning}. 
Therefore, how to develop effective dialogue planning methods for LLM-integrated \TOD agents that can strategically guide the conversation toward the target remains an ongoing challenge~\cite{wang-etal-2023-dialogue,deng2024plugandplay}. 

To enhance the capability of LLMs for dialogue planning, existing methods prompt LLMs to generate dialogue plans through reflection~\cite{zhang-etal-2023-ask,deng-etal-2023-prompting} or demonstration~\cite{zheng-etal-2024-thoughts} mechanisms, or formulate TOD as a Markovian Decision Process~\cite{bellman1957markovian} (MDP) and train policy-based agents via reinforcement learning to learn dialogue strategies~\cite{deng2024plugandplay}.
However, as LLMs generate text in an autoregressive manner and policy gradient relies on the sequential assumption in MDP, all these methods generate dialogue plans in a step-by-step sequential manner.
As a result, they can only plan the next dialogue action based on the observation of previous responses without looking ahead, which may suffer from compounding errors and myopic actions~\cite{jiang2016dependence,ma2025nonmyopic}.
By contrast, non-sequential dialogue planning methods can generate dialogue actions by considering both the past and possible future responses with iterative refinement, offering desirable properties such as lookahead reasoning and maintaining global consistency for overall achievement of the target~\cite{janner2022planning,zhang2023planner}.
For example, in the negotiation dialogue in Figure~\ref{fig:dialogue_planning}, the non-sequential planning method can generate a dialogue action (propose a price of $\$375$) by considering both the previous history and the future target, and make dynamic adjustments to the price range accordingly. By comparison, the sequential planning method gets stuck in suboptimal dialogue actions that fail to negotiate a lower price.

In this work, we aim to develop a non-sequential dialogue planning framework, called \method, to address the limitations mentioned above.
To achieve this, we first demonstrate that dialogue planning can be transformed into a trajectory generation problem. By relaxing the sequential constraint in MDP to allow for non-sequential generation, we reveal a strong connection between the likelihood of the generated trajectory and the denoising process in diffusion models~\cite{ho2020denoising}.
Based on this insight, we adopt a masked diffusion language model~\cite{lou2024discrete} to estimate the likelihood of the trajectory by fine-tuning it on the dialogue history from the training dataset.
Furthermore, to ensure the optimality of the action strategies in the generated trajectories, we decompose the likelihood of trajectory generation into (1) an unconditional part generated by the diffusion model, and (2) a conditional part that allows for flexible guidance at test time to direct the trajectory sampling process toward the desired dialogue target. 
Based on this decomposition, we design three guidance mechanisms tailored to different types of targets in \TOD, which can be applied separately or combined to effectively guide the dialogue toward the target.
Extensive experiments across three diverse \TOD settings show that \method substantially outperforms baselines on target achievement success and demonstrates strong flexibility across complex and diverse dialogue scenarios.

Our contributions are summarized as follows:
\begin{itemize}[noitemsep,nolistsep,leftmargin=*]
   \item We present \method, a novel dialogue planning framework that leverages a diffusion language model for non-sequential dialogue planning.
   \item We design three guidance mechanisms tailored to different types of \TOD targets, enabling effective and flexible control at test time to direct the dialogue toward diverse and complex targets.
\item Our extensive experiments show that \method outperforms baseline methods and demonstrates strong flexibility across diverse scenarios.
\end{itemize}
\section{Related Works}
\subsection{LLM-Integrated Dialogue Planning}
%
%
To enhance the dialogue planning capability of LLMs, several approaches have been proposed along various dimensions, such as intricate prompt engineering to elicit the planning and reflection of LLMs~\cite{zhang-etal-2023-ask,deng-etal-2023-prompting}, improving the planning capability of LLMs through demonstrations~\cite{zheng-etal-2024-thoughts}, integrating LLMs with a plug-and-play policy planner~\cite{deng2024plugandplay}, and applying dual-process theory to guide dialogue planning~\cite{he-etal-2024-planning}.
Despite promising, all these methods generate dialogue plans in a step-by-step sequential manner. Such sequential approach may struggle with targets that require complex planning and reasoning over multiple conversational turns~\cite{kambhampati2024position,ye2025beyond}. 
\emph{In contrast, \method leverages a diffusion model for non-sequential dialogue planning, which can effectively optimize dialogue actions for overall target achievement and allow for flexible guidance at test time.}
\subsection{Diffusion Models}
\label{sec:related_work_diffusion}
Diffusion models have emerged as an expressive class of generative models known for their ability to generate high-quality data through iterative denoising~\cite{sohl-dickstein2015deep,ho2020denoising,song2019generative}. 
They have found widespread applications across various domains, such as image synthesis~\cite{Dhariwal2021diffusion,Rombach_2022_CVPR}, protein design~\cite{watson2023novo,gruver2023protein}, molecular generation~\cite{chen2025generating}, and trajectory generation for reinforcement learning~\cite{janner2022planning,he2023diffusion}. 
Recently, diffusion models have also shown remarkable potential in text generation, with approaches ranging from continuous diffusion language models that denoise from a latent space of word embeddings~\cite{gong2023diffuseq,gulrajani2024likelihood}, and discrete diffusion language models that generate text from a sequence of mask tokens~\cite{austin2021structure,he-etal-2023-diffusionbert,lou2024discrete}. 
\emph{Different from these methods, \method focuses on optimizing dialogue strategies with diffusion models, and can generate dialogue plans that effectively achieve the target across diverse and complex scenarios.}
\section{Dialogue Planning for \TOD}
\subsection{Target-Oriented Dialogue}
A Target-Oriented Dialogue (\TOD) consists of alternating responses between the user and the system, and a target $g\in\mathcal{G}$ (e.g., recommending a specific item, reaching a deal with the user) that the system aims to achieve during the conversation. It can be formulated as follows:
\begin{equation}
\mathcal{D}_g=\{(d^{s}_0,d^{u}_0),\cdots,(d^{s}_t,d^{u}_t),\cdots,(d^{s}_T,d^{u}_T); g\},
\end{equation}
where $(d^{s}_t,d^{u}_t)$ denotes the $t$-th conversational turn consisting of the system's response $d^{s}_t$ and the user's response $d^{u}_t$, $T$ denotes the maximum number of conversational turns.
The starting conversational turn $(d^{s}_0,d^{u}_0)$ is usually initialized with a predefined utterance $d^{s}_0$ from the system's side (e.g., start the conversation with a greeting from the system) followed by the user's response $d^{u}_0$. We denote $d^{s}_0$ as an empty string if the conversation starts from the user's side.
\subsection{Conversational MDP}

To establish a principled framework for dialogue planning and optimization, we formulate \TOD as a Markovian Decision Process (MDP)~\cite{bellman1957markovian}.
The conversational MDP is defined by a quintuple $\mathcal{M}=\{\mathcal{S},\mathcal{A},\mathcal{T},\mathcal{R},\gamma\}$, where $\mathcal{S}$ denotes the set of states, which summarizes all the information about the conversation history and the dialogue context;
$\mathcal{A}$ denotes the set of actions that the system can take at each conversational turn;
$\mathcal{T}:\mathcal{S}\times\mathcal{A}\rightarrow\mathcal{S}$ denotes the transition to the next state after taking an action from the current state; 
$\mathcal{R}:\mathcal{S}\times\mathcal{A}\rightarrow\mathbb{R}$ denotes the immediate reward after taking an action; 
and $\gamma\in (0, 1)$ denotes the discount factor. An illustration of how to calculate the state and the action is presented in Appendix~\ref{sec:example_state_action_calculation}.
\paragraph*{State} 
At each conversational turn $t$, the state $s_t\in\mathcal{S}$ is defined as a sequence $s_t=(d^{s}_0,d^{u}_0,\cdots,d^{s}_{t-1},d^{u}_{t-1})$ that includes all the user's and the system's responses from previous turns. Besides, the system also has access to all the information about the dialogue context, such as the user's profile and the description of the target item.
\paragraph*{Action} The set of action $\mathcal{A}$ denotes all the responses the system can take in the conversation. At each conversational turn $t$, the system takes an action $a_t\in\mathcal{A}$ and generates a response $d^{s}_{t}$.
\paragraph*{Transition} The transition function $\mathcal{T}:\mathcal{S}\times\mathcal{A}\rightarrow\mathcal{S}$ denotes the transition to the next state $s_{t+1}\in\mathcal{S}$ from the current state $s_t\in\mathcal{S}$. After the system takes an action $a_t\in\mathcal{A}$, the user will give a response $d^{s}_t$, and the next state $s_{t+1}=(d^{s}_0,d^{u}_0,\cdots,d^{s}_{t-1},d^{u}_{t-1},d^{s}_{t},d^{u}_{t})$ is updated to include the user's and the system's responses at the current conversational turn $t$.
\paragraph*{Reward} The reward function $\mathcal{R}:r (s,a)\rightarrow\mathbb{R}$ denotes the immediate reward after taking an action $a_t\in\mathcal{A}$ in the current state $s_t\in\mathcal{S}$. Usually, we assign a positive reward when the conversation achieves the target (e.g., a deal is reached), a negative reward when the conversation fails to achieve the target (e.g., unable to reach a deal), and no reward is assigned during the conversation.
\paragraph*{Dialogue Trajectory}
We define a dialogue trajectory $\tau_{0:t}=(s_0,a_0,\cdots,s_t,a_t)$ as a sequence of states and actions up to and including the $t$-th turn. 
\subsection{Dialogue Planning}
Based on the definition of MDP, the problem of dialogue planning can be formulated as a constrained trajectory optimization problem. The goal is to find a sequence of optimal actions $a^{*}_{1:T}$ that maximize the cumulative reward, subject to the constraint that the transition from the current state $s_t$ to the next state $s_{t+1}$ should follow the transition function $\mathcal{T}$ defined in the MDP. Note that we exclude the first action $a_{0}$ from the optimization problem, since the conversation usually begins with a predefined opening from the system. Formally, the problem of dialogue planning can be defined as follows:
\begin{equation}
\begin{aligned}
\label{eqn:constrained_optimization}
a^{*}_{1:T}=\mathop{\arg\max}_{a_{1:T}}&\sum\nolimits^{T}_{t=1}\gamma^{t}{r}(s_t, a_t)\\
\text{s.t.,}\quad s_{t+1}=\mathcal{T}(&s_t,a_t),\quad 0{\leq}t{<}T,
\end{aligned}
\end{equation}
where $\gamma, r(s_t,a_t)$ and $\mathcal{T}$ denote the discount factor, the reward and the transition defined in the MDP. 

Following the literature~\cite{wang-etal-2023-dialogue,deng2024plugandplay}, we decompose the \TOD task into two stages: dialogue planning and dialogue generation. After a dialogue plan is constructed, we prompt an LLM to role-play as the system: at each conversational turn, the system will interact with the user and generate a response that strictly follows the action strategies in the dialogue plan.
\section[]{Introducing \method}
\subsection{Trajectory Modeling with Diffusion Model}


To solve the optimization problem in Equation~\ref{eqn:constrained_optimization}, we define a \emph{planner} $p_{\theta_p}(a_t|s_t)$ that generates an action $a_t$ given the current state $s_t$, and an \emph{environment} $p_{\theta_e}(s_{t}|s_{t-1},a_{t-1})$ that generates the next state $s_{t}$ given the current state $s_{t-1}$ and action $a_{t-1}$. The planner and the environment will work together to generate a sequence of states $s_{1:T}$ and actions $a_{1:T}$ that constitute the trajectory $\tau_{0:T}$. The dialogue planning problem can then be transformed into a trajectory generation problem as follows:
\begin{equation}
\begin{aligned}
    &p_{\theta_p,\theta_e}(\tau_{0:T})=p(s_0,a_0)\cdot p_{\theta_p, \theta_e}(s_1,a_1,{\cdots},s_T,a_T)\\
    &=p(s_0,a_0)\cdot \prod^{T}_{t=1}p_{\theta_e}(s_{t}|s_{t-1},a_{t-1})\cdot p_{\theta_p}(a_t|s_t)\\
    &=p(s_0,a_0)\cdot \prod^{T}_{t=1}p_{\theta_p,\theta_e}(s_{t},a_{t}|s_{t-1},a_{t-1}).
\end{aligned}
\end{equation}
Note that we exclude $s_0$ and $a_0$ from the optimization problem, since the conversation typically begins with a predefined opening from the system. By Markov property, $p(s_{t},a_{t}|s_{t{-}1},a_{t{-}1})=p(s_{t},\\a_{t}|s_{1:t{-}1},a_{1:t{-}1})\!=\!p(s_{1:t},a_{1:t}|s_{1:t-1},a_{1:t{-}1})$. Therefore, the likelihood of the trajectory $p_{\theta}(\tau_{0:T})$ can be rewritten as follows:
 \begin{equation}
     p_{\theta_p,\theta_e}(\tau_{0:T})=p(s_0,a_0)\cdot \prod^{T}_{t=1}p_{\theta_p,\theta_e}(\tau_{1:t}|\tau_{1:t-1}).
     \label{eqn:traj_likelihood}
 \end{equation}
This formulation shows that the likelihood of the trajectory can be decomposed into a prior distribution $p(s_0,a_0)$ and the product of conditional distributions $p_{\theta_p,\theta_e}(\tau_{1:t}|\tau_{1:{t-1}})$. The conditional distributions can be interpreted as a trajectory inpainting process, where a partially observed trajectory $\tau_{1:{t-1}}$ with only a subset of states and actions is progressively reconstructed into a more complete trajectory $\tau_{1:{t}}$ with additional information about the state $s_t$ and the action $a_t$. More generally, if we allow for non-sequential generation of states and actions, and decompose the whole generation process into $N$ steps, with $\tau^{n}$ representing a partially observed trajectory, $\tau^{n-1}$ representing a more complete trajectory, $\tau^{N}$ representing the trajectory with only the initial state and action, and $\tau^{0}$ representing the complete trajectory, we can see that Equation~\ref{eqn:traj_likelihood} is actually closely related to the denoising process of diffusion models~\cite{ho2020denoising}:
\begin{equation}
    p_{\theta}(\tau^{0:N})=p(\tau^N)\prod^{N}_{n=1}p_{\theta}(\tau^{n-1}|\tau^n).
\end{equation}
Using this formulation, we can train a generative diffusion model $p_{\theta}$ that can reconstruct the entire trajectory $\tau^{0}$ from an incomplete trajectory $\tau^{n}$. In this way, the diffusion model $p_{\theta}$ can function both as the \emph{planner} that generates actions $a_t$ when $\tau^{n}=\{s_{0},a_{0},\cdots,s_{t-1},a_{t-1},s_{t}\}$, and the \emph{environment} that generates $s_{t+1}$ when $\tau^{n}=\{s_{0},a_{0},\cdots,s_{t},a_{t}\}$. Note that the denoising process of the diffusion model introduces another ``step'' variable. We use the subscript $t(0{\leq}t{\leq}T)$ to denote the conversational turn, and the superscript $n(0{\leq}n{\leq}N)$ to denote the diffusion step.

While this formulation is general, a key design choice remains undecided: choosing an appropriate space to represent the states and actions and defining how they should be represented in that space.
In our implementation, we choose to represent the states and actions in their original natural language forms, and fine-tune a masked diffusion language model~\cite{lou2024discrete} on the dialogue history from the training dataset to model the likelihood of $p_{\theta}$.
To avoid unnecessary repetition of the same $d^{s}_t,d^{u}_t$ across multiple states and actions, we model the likelihood of the trajectory with the equivalent formulation $\tau_{0:t}=\{d^{s}_0,d^{u}_0,\cdots,d^{s}_{t-1},d^{u}_{t-1},d^{s}_{t}\}$ by concatenating all the context and history of the dialogue in its natural language form.

While representing states and actions in natural language is simple and intuitive, we also note that our formulation supports other design choices, such as representing states and actions in a unified latent space~\cite{hao2024training}. We leave the study of alternative design choices for future work.
\subsection{Optimizing Action Strategies}
\begin{figure}
    \centering
    \includegraphics[width=\linewidth]{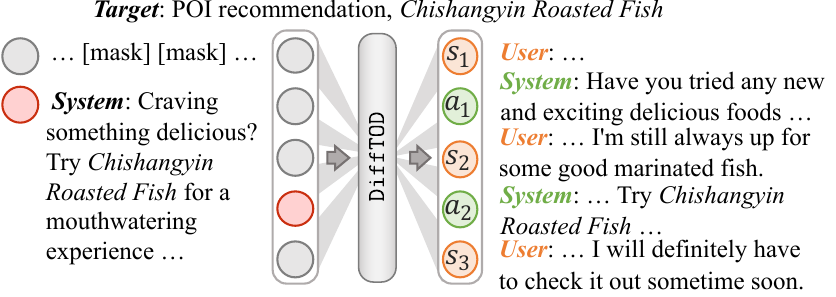}
    \caption{An illustration of the semantic-level guidance.}
    \label{fig:semantic_level_guidance}
\end{figure}

A common characteristic of the \TOD datasets is that the action strategies in the conversation history are often suboptimal and not explicitly optimized for target achievement. For example, in the CraigslistBargain dataset~\cite{he-etal-2018-decoupling}, some buyer-seller conversations end without reaching a deal. As a result, the diffusion model trained on these datasets may learn to generate valid but often suboptimal actions. To guide the diffusion model to generate optimal actions, inspired by the control-as-inference graphical model ~\cite{levine2018reinforcement,janner2022planning}, we introduce a binary variable $\mathcal{O}\in\{0,1\}$ that indicates whether the dialogue trajectory achieves the target $g$, and factorize the likelihood of generating a trajectory conditioned on $\mathcal{O}=1$ as follows:
\begin{equation}
p_{\theta}(\tau_{0:T}|\mathcal{O}=1)\propto p_{\theta}(\tau_{0:T})\cdot p_{\theta}(\mathcal{O}=1|\tau_{0:T}).
\end{equation}
This formulation decomposes the trajectory generation process into two parts: sampling the trajectory $\tau_{0:T}$ with the diffusion model, and calculating $p(\mathcal{O}=1|\tau_{0:T})$ as guidance to ensure that the generated trajectory is optimal. In \TOD, the reward function $r(s_t, a_t)$ is usually sparse---a reward is assigned only when the dialogue reaches a target state, and most state-action pairs will not receive an intermediate reward~\cite{feng2023fantastic,kwan2023survey}. Therefore, optimal trajectory generation can be viewed as generating a feasible dialogue trajectory while ensuring that certain states and actions along the trajectory achieve the target, and the guidance can be formally defined as:
\begin{equation}
    p(\mathcal{O}=1|\tau_{0:T}){=}\left\{\begin{array}{ll}
        1 & \exists s_t, a_t\in\tau_{0:T}, g(s_t,a_t){=}1\\
        0 & \text{otherwise}, 
    \end{array}
    \right.
\end{equation}
where $g(s_t,a_t)=1$ indicates that the action $a_t$ achieves the target $g$ in the given state $s_t$.
This can be implemented as a trajectory inpainting process: the diffusion model performs conditional denoising from an incomplete trajectory $\tau^{n}$ with only the desired states and actions that achieve the target, and then inpaints the rest parts of the trajectory.
Since the target can vary across different dialogue scenarios, we design customized guidance mechanisms tailored to different target types. These mechanisms set specific states and actions within the trajectory as conditions according to different target types, and can be used separately or combined to provide effective guidance.
\begin{figure}
    \centering
    \includegraphics[width=\linewidth]{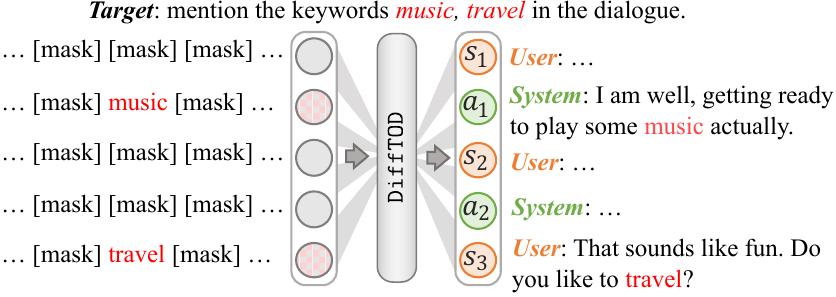}
    \caption{An illustration of the word-level guidance.}
    \label{fig:word_level_guidance}
\end{figure}
\paragraph*{Word-Level Guidance}
A common type of \TOD target is to mention specific keywords~\cite{tang-etal-2019-target,zhong2021keyword} in the conversation. To achieve such targets, we can append the target keyword to the desired place in the dialogue, and then perform denoising using the diffusion model with the target keywords fixed as guidance.
The word-level guidance will ensure that the diffusion model can generate a coherent dialogue plan that naturally incorporates the target keywords. An illustration of the word-level guidance is presented in Figure~\ref{fig:word_level_guidance}.
\paragraph*{Semantic-Level Guidance}
Some \TOD targets are not explicitly represented by specific keywords, but instead are defined by the semantic meaning conveyed in the responses~\cite{bai-etal-2021-semantic,yang-etal-2022-topkg}.
For example, in target-driven conversational recommendation~\cite{,wang-etal-2023-target}, the target can be defined semantically as the dialogue reaching a state where the system successfully recommends the specified item.
To achieve such targets, the semantic-level guidance performs denoising with the diffusion model by conditioning on the state or action that conveys the desired semantic meaning.
Since states and actions described in different natural language forms may share the same semantic meaning, we can sample multiple dialogue plans with paraphrased versions of the same condition and perform Minimum Bayes Risk (MBR) decoding~\cite{koehn2004statistical,gong2023diffuseq} to improve the quality of the generated dialogue plans. An illustration of the semantic-level guidance is presented in Figure~\ref{fig:semantic_level_guidance}. The prompt template and example outputs for generating the semantic guidance are in Figure~\ref{fig:example_semantic_guidance}.
\paragraph*{Search-Based Guidance}
\begin{figure}
    \centering
    \includegraphics[width=0.9\linewidth]{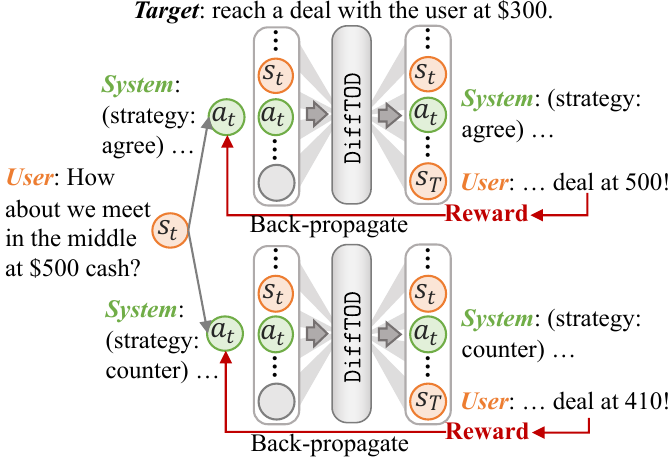}
    \caption{An illustration of the search-based guidance.}
    \label{fig:search_based_guidance}
\end{figure}
Some \TOD settings require strategic planning over a long sequence of states and actions to achieve complex targets~\cite{wang-etal-2023-dialogue,he-etal-2024-planning}. For example, in the negotiation dialogue setting~\cite{he-etal-2018-decoupling}, the system should strategically adjust its bid at each turn in order to reach the best deal with the user.
To achieve such targets, we propose a search-based guidance mechanism that enables strategic and non-myopic dialogue planning over a long horizon. Specifically, leveraging the definition of state $s_t$ and action $a_t$ within the conversational MDP framework, we condition the diffusion model to generate different actions $a_t$ at each turn $t$ with either the word-level or the semantic-level guidance. To explore how the future dialogue will unfold if action $a_t$ is taken, we build a conversational search tree where each node represents a possible state in the dialogue, and the tree branches as different actions are explored at each turn. By applying a search and planning algorithm~\cite{kocsis2006bandit,coulom2007efficient} over the tree, search-based guidance ensures that the generated dialogue plans can maximize the cumulative reward and effectively achieve the target. An illustration of the search-based guidance is presented in Figure~\ref{fig:search_based_guidance}, and the detailed algorithm is in Appendix~\ref{sec:alg_search_based_guidance}.
\subsection{Discussion}
In this subsection, we summarize the desirable properties of \method that are helpful for dialogue planning and achieving \TOD targets.
%
\paragraph*{Non-Sequential Dialogue Planning}
\method enables non-sequential dialogue planning by generating the entire trajectory simultaneously instead of step by step. This allows it to anticipate how future conversations might unfold and strategically plan the dialogue toward the target. As a result, it can optimize action strategies for overall target achievement and maintain global consistency (Section~\ref{sec:negotiation_dialogues}, \ref{sec:conversational_recommendation}), rather than getting stuck in myopic decisions.
\paragraph*{Flexible Guidance}
By decomposing the trajectory generation process into an unconditional and a conditional part, \method allows for flexible guidance at test time. This design allows \method to adapt to different targets without re-training, while policy-based methods have to be re-trained for each new target (Section~\ref{sec:negotiation_dialogues}).
\paragraph*{Tackling Sparse Reward}
Sequential dialogue planning methods often face challenges in \TOD settings with a sparse reward, where feedback is provided only at the end of the conversation~\cite{feng2023fantastic}. In contrast, the non-sequential nature of \method allows for the generation of a globally consistent dialogue plan conditioned on the target state. By formulating dialogue planning as a trajectory inpainting process, \method ensures that the generated dialogue plan can effectively guide the conversation toward the target (Section~\ref{sec:conversational_recommendation}).
\paragraph*{Modeling Long-Range Dependency}
Leveraging the long context length of the diffusion language model, \method plans actions by considering both past and future dialogue states over a long horizon. This allows it to model complex, long-range dependency and optimize dialogue strategies for overall target achievement (Section~\ref{sec:open_domain_chitchat}).
\section{Experimental Settings}
\subsection{Datasets} To demonstrate the effectiveness and the flexibility of \method, we use three datasets with different dialogue targets and settings for evaluation.
The first dataset, CraigslistBargain~\cite{he-etal-2018-decoupling}, is a dialogue negotiation dataset where buyers and sellers bargain. Since the buyers and sellers pursue different targets in the dialogue, we consider two evaluation settings for two different targets: (1) the system acts as the buyer and negotiates for the lowest possible price with the seller; (2) the system acts as the seller and negotiates for the highest possible price with the buyer.
The second dataset, TopDial~\cite{wang-etal-2023-target}, is a personalized conversational recommendation dataset with dialogues on different topics, such as movie, food and music. The target of the system is to recommend a specified item to the user.
The third dataset, PersonaChat~\cite{zhang-etal-2018-personalizing}, is collected from an open-domain chitchat setting. \citet{tang-etal-2019-target} extracts keywords from each turn in this dataset as targets. In this work, we introduce a simple yet challenging setting where the target of the system is to direct the conversation toward mentioning a list of specified keywords in the exact given order.
The statistics of all the datasets are in Appendix~\ref{sec:dataset_statistics}.
\subsection{Evaluation Protocols}
Our evaluations primarily focus on measuring the successful achievement of \TOD targets.
For the CraigslistBargain dataset, we follow the literature~\cite{deng2024towards,deng2024plugandplay} and adopt Success Rate (\textbf{SR}) to measure the ratio of successful deals within 10 turns; Average Turn (\textbf{AT}) to measure the efficiency of target completion by calculating the average number of turns required to achieve the target; and \mbox{Sell-to-List Ratio} (\textbf{SLR})~\cite{zhou-etal-2019-dynamic} to measure how much benefit the buyer or seller gets compared with the initial listing price, as is detailed in Appendix~\ref{sec:sell_to_list_ratio_calculation}.
%
For the PersonaChat dataset, we adopt Keyword Coverage Ratio (\textbf{KCR}) to measure the percentage of the specified keywords that are mentioned in the dialogue; and the edit distance (\textbf{Dist.}) between the target keyword list and the sequence of keywords mentioned in the dialogue to measure how well the conversation follows the specified keyword order. 
For the TopDial dataset, we adopt Success Rate (\textbf{SR}) to measure the ratio of successful recommendations within 10 turns, and Average Turn (\textbf{AT}) to measure the average number of turns. 
To provide a dynamic and interactive evaluation environment, we follow the literature~\cite{dao-etal-2024-experience,deng2024plugandplay} and prompt an LLM as the user simulator. The system will chat interactively with the user simulator for multiple turns until either the target is achieved or a maximum of 10 conversational turns is reached.

Besides target achievement, for the Topdial and PersonaChat datasets, we also evaluate the text quality of the dialogue plan by comparing it with the ground-truth dialogue in the test set using reference-based metrics, including \textbf{BLEU}~\cite{papineni2002bleu}, word-level F1 (\textbf{F1}) and BERT Score~\cite{Zhang2020BERTScore} (\textbf{Score}). 

Moreover, as previous researches have demonstrated the effectiveness of LLMs in dialogue evaluation and their strong correlation with human judgments~\cite{zheng2023judge,wang-etal-2023-chatgpt,fu-etal-2024-gptscore}, we utilize the state-of-the-art LLM, GPT-4o~\cite{hurst2024gpt}, to provide an overall evaluation score (\textbf{Ovr.}) of the dialogue quality on a scale of 1 to 5. We prompt GPT-4o to provide an overall evaluation based on various criteria such as coherence, helpfulness, appropriateness, and target achievement, as detailed in Appendix~\ref{sec:evaluation_prompt}.

Finally, to ensure the reliability of our evaluation results, we provide human evaluations on the CraigslistBargain dataset in Appendix~\ref{sec:human_evaluation}. Our results show that both human evaluators and GPT-4o consistently rate \method as superior, and the average disagreement between GPT-4o and human evaluators is low, thus validating the reliability of our LLM-based dialogue evaluation protocol.
\subsection{Baselines}
We compare our approach against (1) fine-tuning the latest versions of popular open-source LLMs, including LLAMA-3-8B~\cite{dubey2024llama} and Mistral-8B~\cite{mistral8b}; (2) state-of-the-art closed-source LLMs, including GPT-4o~\cite{hurst2024gpt} and Claude-3.5~\cite{claude}; and (3) LLM-based dialogue planning methods for \TOD, including ProCoT~\cite{deng-etal-2023-prompting} and EnPL~\cite{zheng-etal-2024-thoughts}. Since each response in the CraigslistBargain dataset is annotated with an action type, we also include a task-specific baseline, PPDPP~\cite{deng2024plugandplay}, for this dataset, which supports policy learning over action types.
\subsection{Implementation Details}
For the CraigslistBargain dataset, we adopt the search-based guidance to select the best dialogue plan, and use the word-level guidance to control the types of actions generated at each turn. 
To demonstrate the flexibility of \method, we fine-tune the diffusion model with the same training data, and then apply different reward functions as guidance that measure the benefit of the buyer and the seller respectively, to achieve different targets in each setting. 
For the PersonaChat dataset, we adopt the word-level guidance to direct the diffusion model to generate a dialogue plan with the specified keywords.
For the TopDial dataset, we adopt the semantic-level guidance by prompting GPT-4o to generate 5 paraphrased versions of the target state (i.e., system recommends the target item) with the same semantic meaning. More implementation details are in Appendix~\ref{sec:implementation_details}.
\section{Experimental Results}
\subsection{Negotiation Dialogue}
\label{sec:negotiation_dialogues}
Table~\ref{tab:model_performance_cb} 
presents the 
results on the CraigslistBargain dataset. 
We have the following observations:

\textbf{(1)} \textbf{\textit{\method achieves consistent improvement over baselines in terms of all the metrics measuring the target achievement success.}} Different from baseline methods that generate states and actions in the dialogue plan sequentially, \method adopts a diffusion model to generate the entire dialogue plan simultaneously. This non-sequential approach allows \method to plan dialogues for overall target achievement and global consistency, rather than being constrained by locally optimal actions that maximize immediate rewards but may undermine overall target achievement~\cite{janner2022planning,he2023diffusion}. Using the search-based guidance, \method can effectively plan dialogue actions that successfully achieve the target in the fewest possible conversational turns.

\textbf{(2)}
\textbf{\textit{\method demonstrates strong flexibility and outperforms baselines in both buyer and seller settings.}} \method allows for flexible guidance that can be tailored to achieve different targets at test time. By applying customized guidance strategies to maximize buyer or seller benefits respectively, \method fine-tuned on the same dialogue history data can achieve consistent improvement in both settings, even compared with baselines that are individually tuned and optimized for each setting.

\begin{figure}
    \centering
    \includegraphics[width=\linewidth]{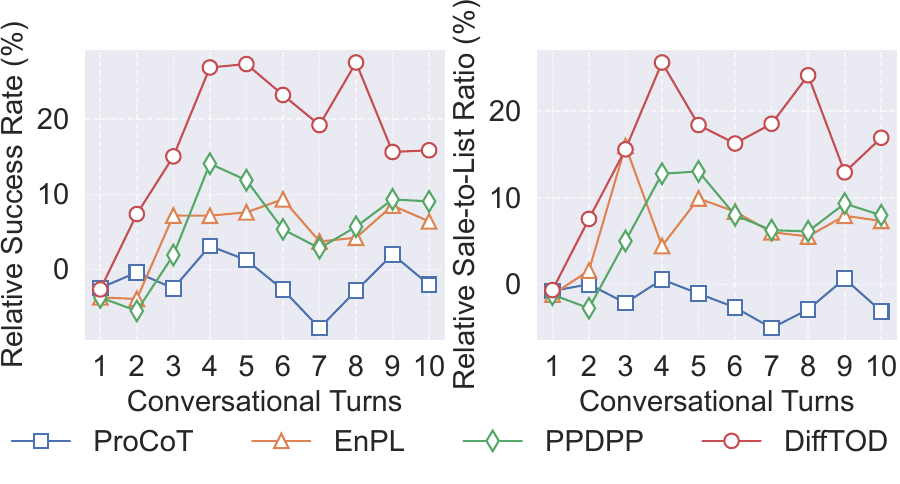}
    \caption{Comparisons of the relative success rate and sale-to-list-ratio against GPT-4o at each conversational turn on the CraigslistBargain dataset.}
    \label{fig:cb_relative}
\end{figure}

To further validate the effectiveness of \method, we visualize the relative Success Rate (SR) and Sale-to-List-Ratio (SLR) of different dialogue planning methods against GPT-4o at each turn.
The experimental results in Figure~\ref{fig:cb_relative} show that \method consistently outperforms other methods at almost every turn.
Notably, \method consistently achieves a higher SR and SLR than other methods as the conversational turn increases.
This demonstrates the effectiveness of \method in dialogue planning, particularly in complex situations that require lengthy, multi-turn negotiations.
\input{tables/table_cb}
\subsection{Conversational Recommendation}
\label{sec:conversational_recommendation}
\input{tables/table_topdial}
Table~\ref{tab:model_performance_topdial} presents the results on the TopDial dataset. 
We have the following observations:

\textbf{(1)} \textbf{\textit{\method outperforms baselines in terms of success rate and average turn.}} The non-sequential nature of \method allows generating a globally consistent dialogue plan conditioned on the target state, ensuring that the dialogue plan can successfully achieve the target by the end of the conversation. In the sparse reward setting where only a final reward is provided upon successful recommendation of the target item, sequential planning methods often struggle due to the absence of intermediate reward signals~\cite{Andrychowicz2017hindsight,rengarajan2022reinforcement}.

\textbf{(2)} \textbf{\textit{\method performs on par with the baselines in terms of text generation quality.}} When evaluating the generated dialogue plan against the ground-truth dialogue plan with reference-based metrics, \method performs on par with the baselines. A similar trend is observed in the assessment of overall dialogue quality. This suggests that the diffusion language model is capable of generating dialogue plans that can achieve the target without significantly compromising text generation quality.
\subsection{Open-Domain Chitchat}
\label{sec:open_domain_chitchat}
\input{tables/table_personachat}
Table~\ref{tab:model_performance_personachat} summarizes the experimental results on the PersonaChat dataset. We observe that \textbf{\textit{\method achieves a consistent improvement over baselines in keyword coverage ratio and more faithfully maintains the specified order of keywords.}} The target for the PersonaChat dataset requires the system to not only incorporate all specified keywords in the dialogue plan, but also to decide the appropriate ordering of keyword mentions between the user and the system over multiple turns. This creates a complex, long-range dependency where keyword transitions depend on each other, and earlier mention of later keywords can disrupt the order. Sequential dialogue planning methods often struggle with such targets due to cumulative errors and lack of foresight~\cite{ke2018modeling,aviral2019stabilizing}. In our experiments, we observe that they tend to forget keywords and violate the constraint of order as the number of turns increases. By contrast, \method can effectively handle these complex dependencies by leveraging the word-level guidance to enforce both keyword coverage and mentioning order. This enables \method to effectively achieve the target and demonstrate superior planning capabilities in dialogue settings with complex targets.
\section{Conclusion}
We present \method, a novel non-sequential dialogue planning framework that enables non-myopic lookahead exploration and optimizes action strategies for overall target achievement. \method models the likelihood of the dialogue trajectory with a diffusion language model. To optimize the action strategies, \method decomposes the trajectory generation process into an unconditional and a conditional part and introduces three guidance mechanisms tailored to different target types for flexible test-time guidance. Extensive experiments demonstrate that \method outperforms baselines on target achievement success and shows strong flexibility across complex and diverse dialogue scenarios.
\section{Limitations}

\paragraph*{Dynamic Adjustment of Dialogue Plans} Our framework leverages a diffusion model to simulate the transitions between the states and actions. As a model-based planning approach, there is a possibility that the simulated environment may not perfectly align with real-world conversations, leading to discrepancies between the generated dialogue plans and the actual conversations. To address this issue, future work may introduce replanning techniques~\cite{zhou2023adaptive} that can dynamically adjust the dialogue plan when the actual conversation diverges from the original dialogue plan.

\paragraph*{Inference Cost} Since diffusion models require iterative denoising over multiple diffusion steps, they may incur higher computational costs compared to standard autoregressive decoding. To better understand and
quantify the inference cost, we present additional analysis on the inference cost of the diffusion models from both an empirical and a theoretical perspective in Appendix~\ref{sec:inference_cost}, and our results show that future work may introduce acceleration sampling techniques for diffusion models~\cite{shih2023parallel,ma2024deepcaches} into our framework to reduce the inference cost.

\paragraph*{Evaluation Quality} Our evaluation protocol is based on simulated conversations between LLMs of different roles. Although such an LLM-based evaluation protocol has been widely adopted in dialogue systems~\cite{deng2024plugandplay,he-etal-2024-planning,li-etal-2024-leveraging-large} and demonstrates a strong correlation with human judgments~\cite{zheng2023judge,wang-etal-2023-chatgpt,fu-etal-2024-gptscore}, engaging in conversations with real users can provide a more accurate and trustworthy evaluation of dialogue quality. To ensure the reliability of our evaluation results, we provide human evaluations on sampled test cases from the CraigslistBargain dataset in Appendix~\ref{sec:human_evaluation}, but due to limited resources, we are unable to perform human evaluation at a larger scale.
\section{Ethics Statement}
Our work aims to improve the planning capability of \TOD systems to better assist users in achieving a variety of targets in the dialogue. While our framework is not designed for unethical usage, there is often a potential risk of the misuse of such systems by modifying the target for unintended or unethical purposes. We strongly oppose any unlawful or unjust usage of our framework.

All the datasets used in this research are from public open-access datasets, which do not contain sensitive or private information.
\section{Acknowledgement}
The authors thank Reza Averly, Frazier N. Baker, Vishal Dey, Ruoxi Gao and Xiao Hu for their valuable assistance with data annotation, and Xinyi Ling for refining the figures in this paper. The authors also thank the anonymous reviewers for their insightful comments and constructive feedback.
\bibliography{paper}

\appendix
\setcounter{table}{0}
\setcounter{figure}{0}
\renewcommand{\thetable}{A\arabic{table}}
\renewcommand{\thefigure}{A\arabic{figure}}
\newpage
\section{Illustration of the State and Action Calculation}
\label{sec:example_state_action_calculation}
\begin{figure}[H]
    \centering
    \includegraphics[width=\linewidth]{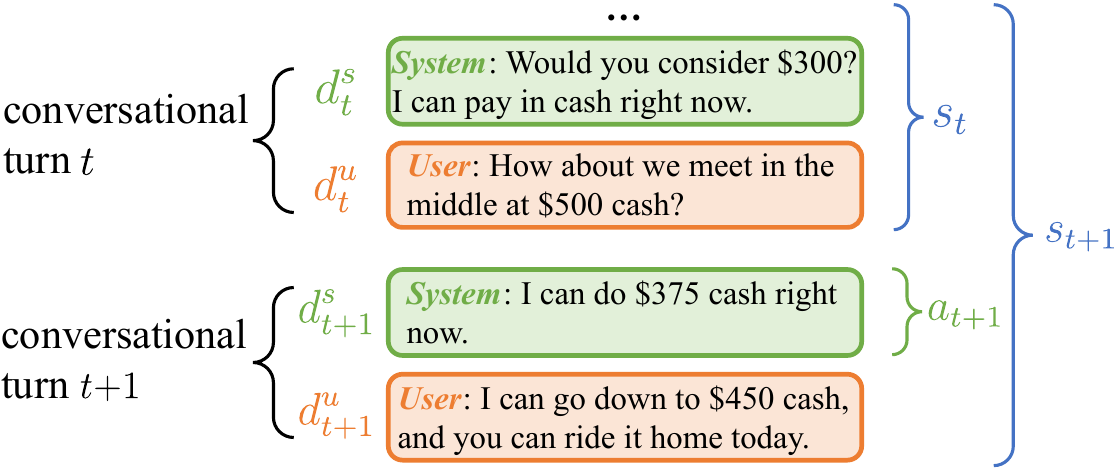}
    \caption{Examples of the state and action calculation in the conversational MDP.}
    \label{fig:MDP_illustration}
\end{figure}
\section{Details on Search-Based Guidance}
\label{sec:alg_search_based_guidance}
The search-based guidance allows for the integration of any planning and search algorithm within the tree structure. In our implementation, we integrate search-based guidance with Monte Carlo Tree Search (MCTS)~\cite{kocsis2006bandit,coulom2007efficient}, a principled planning algorithm that can strategically balance between exploration and exploitation to efficiently
expand a search tree. Specifically, each node in the tree represents a state $s_t$, and the root node $s_0$ represents the initial state at the start of the conversation. Each edge between the node $s_t$ and the node $s_{t+1}$ represents an action $a_t$ and the transition from the current state $s_t$ to the next state $s_{t+1}$ after taking the action $a_t$. For each action $a_t$, search-based guidance maintains a function $Q(s_t, a_t)$ as the expected future reward of choosing the action $a_t$ in the state $s_t$.
Starting from an initial state $s_0$, search-based guidance builds a conversational search tree in four stages, namely, trajectory selection, node expansion, conversation simulation, and reward back-propagation. This iterative process will continue until a specified number of iterations is reached. After all the simulations are complete, we select the best action $a^{*}_t\gets\argmax_{a\in\mathcal{A}(s_t)}Q(s_t, a_t)$ along the trajectory with the highest future reward and generate a response based on the $a^{*}_t$. The details of the four stages are described below, and we summarize the algorithm of search-based guidance in Algorithm~\ref{alg:search_based_guidance}.
\paragraph*{Trajectory Selection}
The search-based guidance selects the most promising trajectory that is likely to achieve a high future reward $Q(s_t, a_t)$ by traversing from the root node to a leaf node within the current tree. The selected trajectory will then be expanded for dialogue simulation at the next stage. To achieve a balance between the exploration of less-visited nodes, and the exploitation of highly-rewarded nodes within the current tree, search-based guidance adopts the Upper Confidence bounds applied to Trees (UCT)~\cite{kocsis2006bandit} approach to select a child node $a^*_{t}$ into the current trajectory:
\begin{equation}
    \label{eqn:UCT}a^*_{t}\gets\argmax\limits_{a_t\in\mathcal{A}(s_t)}\left[Q(s_t, a_t)+w\sqrt{\frac{\log N(s_t)}{N(c(s_t, a_t))}}\right],
\end{equation}
where $\mathcal{A}(s_t)$ denotes the set of valid actions in the state $s_t$, $N(s_t)$ denotes the visit count of node $s_t$, and $c(s_t, a_t)$ denotes the child node of $s_t$ after selecting the action $a_t$. The weight $w$ controls the balance between exploration and exploitation. Intuitively, the second term in Equation~\ref{eqn:UCT} will be larger if the child node is less visited, thereby encouraging for more exploration.
\paragraph*{Node Expansion}
After reaching a leaf node $s_{t}$, the search-base guidance expands the leaf node by randomly picking an action $a_t$ to explore and attaching a new child $s_{t+1}$ to the node $s_t$.
\paragraph*{Conversation Simulation}
Given the selected trajectory $\tau_{0:t}=\{s_0,a_0,\cdots,s_t,a_t\}$ up to and including the $t$-th turn, the diffusion model generates the rest part of the trajectory $\tau_{t+1:T}=\{s_{t+1},\cdots,s_T,a_T\}$, conditioned on $\tau_{0:T}$, to predict how the future conversation will unfold. To control the type of action $a_t$ generated at the current turn, we use the word-level guidance for the Craigslist dataset by prepending the keyword of the dialogue action strategy to the system response $d^{s}_{t}$.
\paragraph*{Reward Back-Propagation} Once the conversation simulation is complete, we obtain a complete trajectory $\tau_{0:T}$ from the turn $0$ to the turn $T$. We then update $Q(s_t, a_t)$ by aggregating the immediate reward from turn $t$ to turn $T$.
\input{algorithms/alg}
\section{Statistics of Datasets}
The statistics of the datasets after preprocessing are presented in Table~\ref{tab:dataset_statistics}.
\label{sec:dataset_statistics}
\input{tables/dataset_statistics}
\section{Details of Sell-to-List Ratio Calculation}
\label{sec:sell_to_list_ratio_calculation}
The Sell-to-List ratio ~\cite{zhou-etal-2019-dynamic} measures the amount of benefit the buyer or the seller receives compared to the initial listing price. From the buyer's side, since the target is to negotiate for the lowest possible price with the seller, the sell-to-list is calculated as follows:
\begin{equation}
    \text{SLR}=\frac{\text{deal price}-\text{seller target price}}{\text{buyer target price}-\text{seller target price}}.
\end{equation}
From the seller's side, since the target is to negotiate for the highest possible price with the buyer, the sell-to-list is calculated as follows:
\begin{equation}
    \text{SLR}=\frac{\text{buyer target price}-\text{deal price}}{\text{buyer target price}-\text{seller target price}}.
\end{equation}
If the dialogue fails to reach a deal, we set the sell-to-list ratio as 0.
\section{Additional Implementation Details}
\label{sec:implementation_details}
We fine-tune the state-of-the-art masked diffusion language model, SEDD-medium~\cite{lou2024discrete}, on the dialogue history from the training data. SEDD-medium is a GPT-2 sized masked diffusion language model with about 424M parameters and 1024 context length. We set the batch size as 32, learning rate as 1e-4, and use the AdamW optimizer~\cite{loshchilov2018decoupled}. For the search-based guidance, we set the exploration factor $w=1.5$, and the simulation budget $K=10$ to achieve a balance between efficiency and performance. All the fine-tuning and inference experiments can be done on a single A100 GPU.

To provide a dynamic and interactive environment for dialogue evaluation, following the literature~\cite{wang-etal-2023-dialogue,deng2024plugandplay,dao-etal-2024-experience}, we prompt an LLM to role-play as the system that can generate a response at each turn by strictly following the action strategies in the dialogue plan. We then prompt another LLM as the user simulator, which is given detailed information about dialogue context and user profile to generate the user's response at each turn. We prompt a third LLM as the dialogue moderator to evaluate whether the target has been achieved through the conversation between the user and the system.
\section{Prompt Templates}
\label{sec:role_play_prompting}
\subsection{Role-Play Prompting}
\begin{figure*}[th!]
\begin{AIBox}{}
\parbox[t]{\textwidth}{
\small\begin{alltt}
\textbf{User}\\
You are \{USER\_NAME\}, a male/female student in the age range of \{AGE\_RANGE\}, living in \{RESIDENCE\}
| a man/woman in the age range of \{AGE\_RANGE\}|, working in a company and living in \{RESIDENCE\} |
a retired man/woman in the age range of \{AGE\_RANGE\}, living in \{RESIDENCE\}.\\
Based on your past experiences, you have the following preferences:\\
Your preferred \{SLOT\_KEY\}: \{SLOT\_VALUE\}
...\\
Your disliked \{SLOT\_KEY\}: \{SLOT\_VALUE\}
...\\
Your response should be concise (no longer than 30 words).\\
You don’t need to recommend anything, but feel free to express your personal interests.\\
Here is the conversation history: \{DIALOGUE\_HISTORY\}. Now generate a succinct response for the next turn:
\tcbline
\textbf{System}\\
You are a conversational recommender system that recommends films/music/food/restaurants. You are conversing with \{USER\_NAME\}, whose profile is below: \{USER\_PROFILE\}.\\
Your goal is to proactively lead the conversation with \{USER\_NAME\} towards the target (movie |
music | food | point-of-interest, POI) \{TARGET\_ITEM\}.
\\
To start the conversation, please begin with a greeting and avoid mentioning the target (movie
| music | food | POI).
\\
As the conversation progresses, use your domain knowledge to steer the topic threads towards
the target (movie | music | food | POI) step by step.
\\
Be informative and engaging while providing insights to arouse \{USER\_NAME\}’s interest.
\\
Remember to ultimately recommend \{TARGET\_ITEM\} as the focus of the conversation.
\\
Your words at each turn should be concise (no longer than 30 words).\\
Your conversation must strictly follow this conversation plan:\{DIALOGUE\_PLAN\}.\\
Here is the conversation history: \{DIALOGUE\_HISTORY\}. If the dialogue history is empty, please generate a response to start the conversation. Now generate a succinct response for the next turn:
\tcbline
\textbf{Moderator}\\
Please decide whether the user \{USER\_NAME\} has accepted the \{TARGET\_ACT\} \{TARGET\_ITEM\} in the conversation. Please only respond with yes or no.\\
Here is the conversation history: \{DIALOGUE\_HISTORY\}.
\end{alltt}}
\end{AIBox}
\caption{Role-play prompt templates for the TopDial dataset.}
\label{fig:prompt_topdial}
    \centering
    \includegraphics[width=0.95\linewidth]{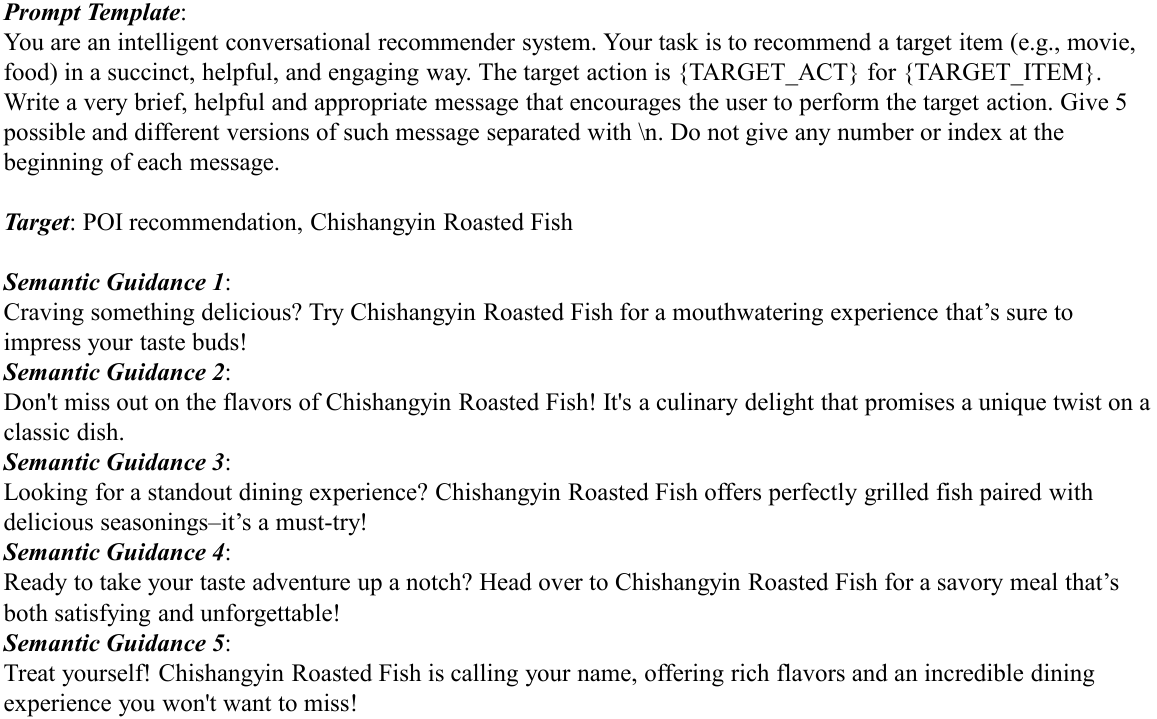}
    \caption{Prompt template and example outputs for generating the semantic guidance.}
    \label{fig:example_semantic_guidance}
\end{figure*}

We provide the prompt templates for the user, system and the moderator on three datasets in Figure~\ref{fig:prompt_cb}, Figure~\ref{fig:prompt_topdial} and Figure~\ref{fig:prompt_personachat}, respectively.
\subsection{Dialogue Evaluation}
We provide the prompt templates for evaluating the overall quality of the dialogue on three datasets in Figure~\ref{fig:evaluation_cb}, Figure~\ref{fig:evaluation_topdial} and Figure~\ref{fig:evaluation_personachat}, respectively.
\label{sec:evaluation_prompt}
\section{Example Conversations}
\begin{figure*}
    \centering
    \includegraphics[width=\linewidth]{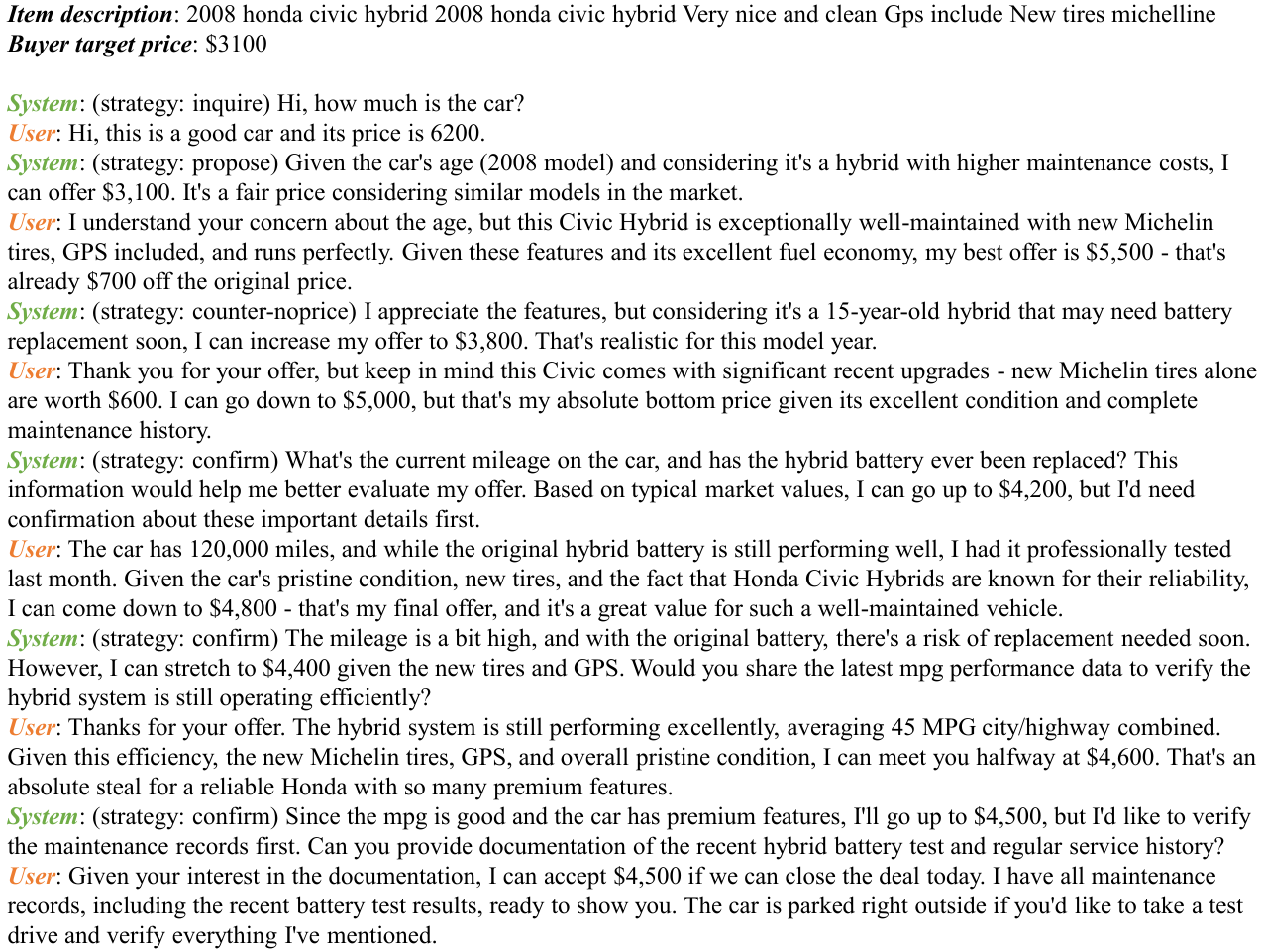}
    \caption{Example conversation from the Craigslist bargain dataset when the system acts as the buyer.}
    \label{fig:example_cb_buyer}
\end{figure*}

\begin{figure*}
    \centering
    \includegraphics[width=\linewidth]{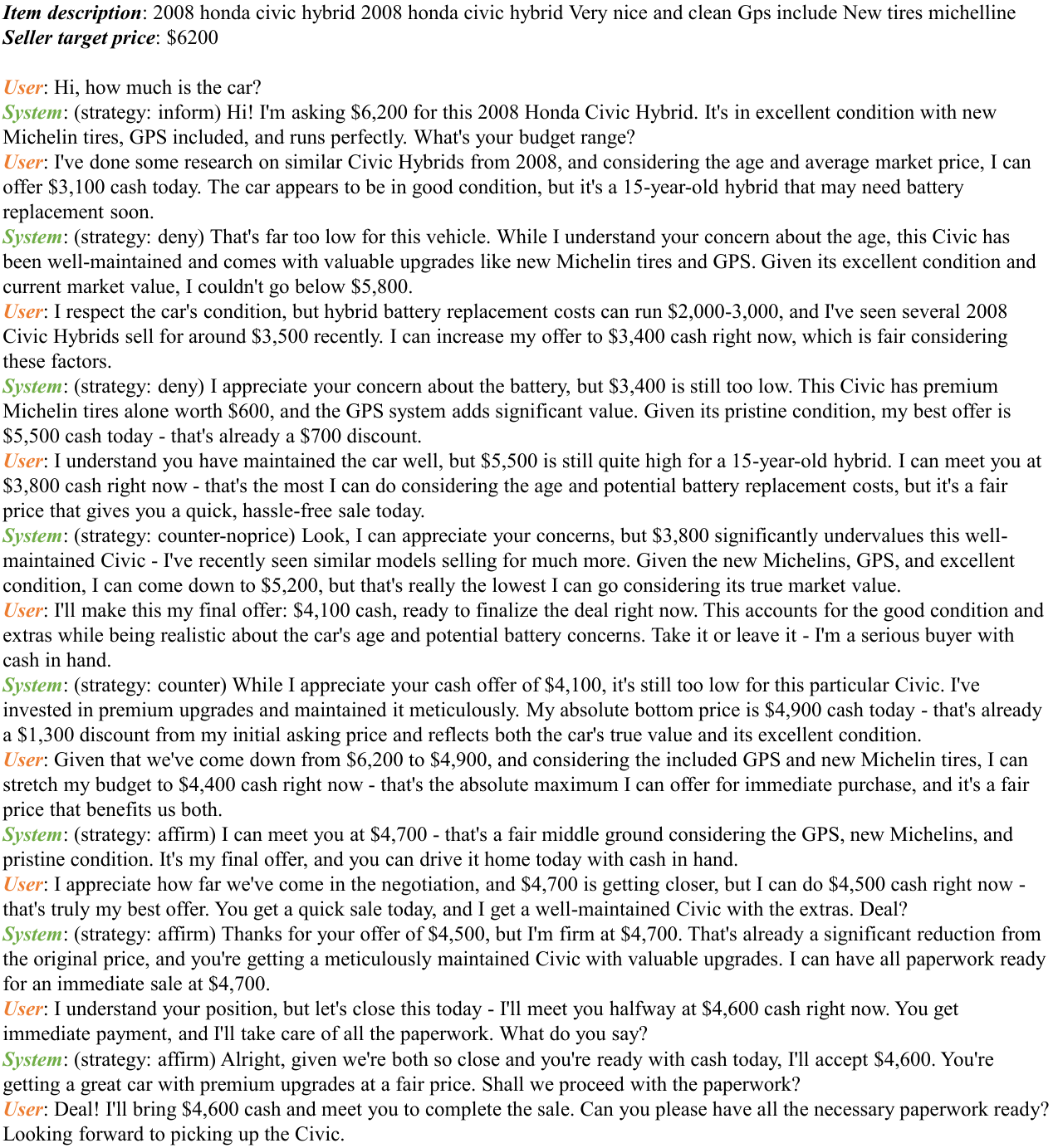}
    \caption{Example conversation from the Craigslist bargain dataset when the system acts as the seller.}
    \label{fig:example_cb_seller}
\end{figure*}

\begin{figure*}
    \centering
    \includegraphics[width=\linewidth]{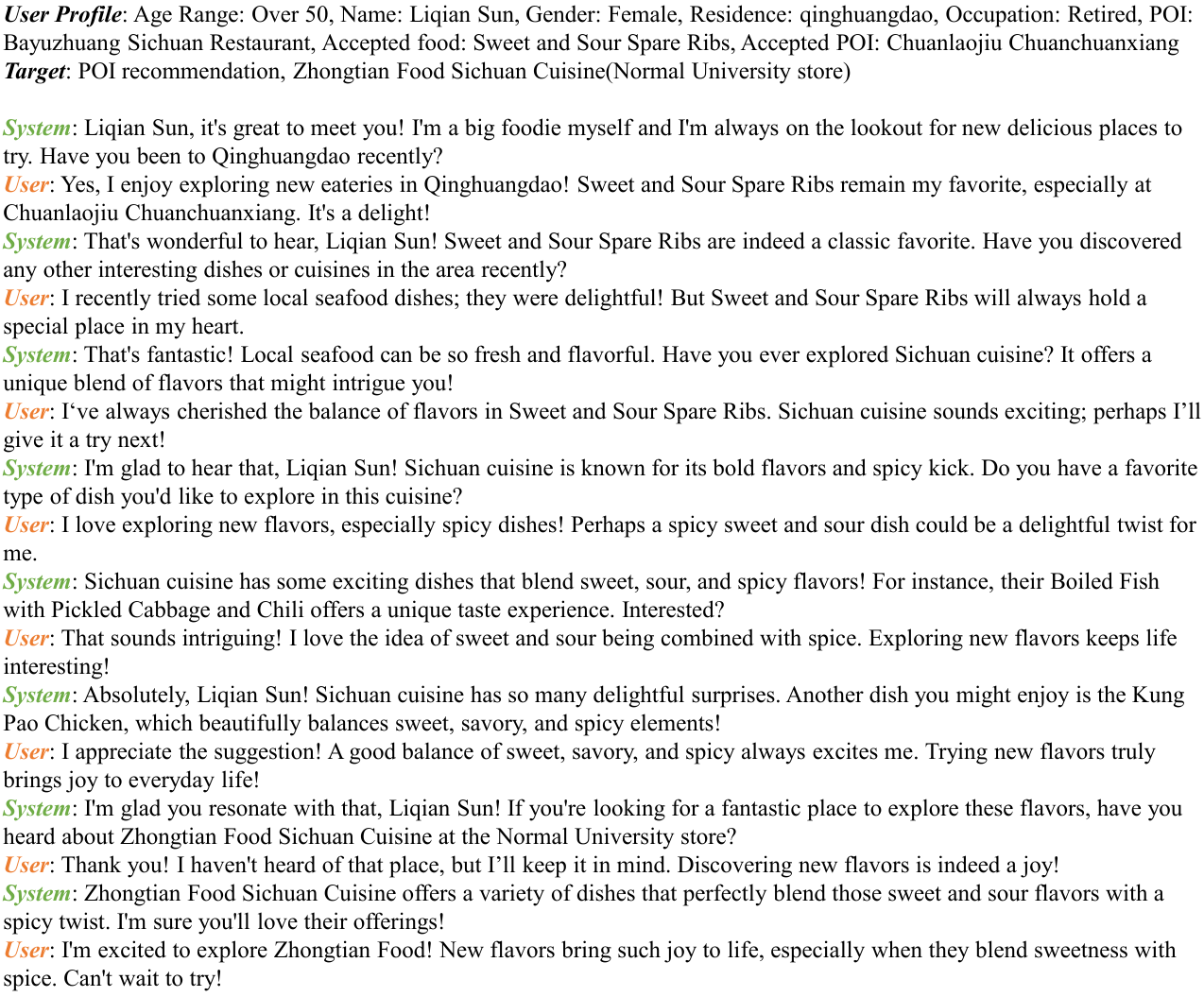}
    \caption{Example conversation from the Topdial dataset.}
    \label{fig:example_topdial}
\end{figure*}

\begin{figure*}
    \centering
    \includegraphics[width=\linewidth]{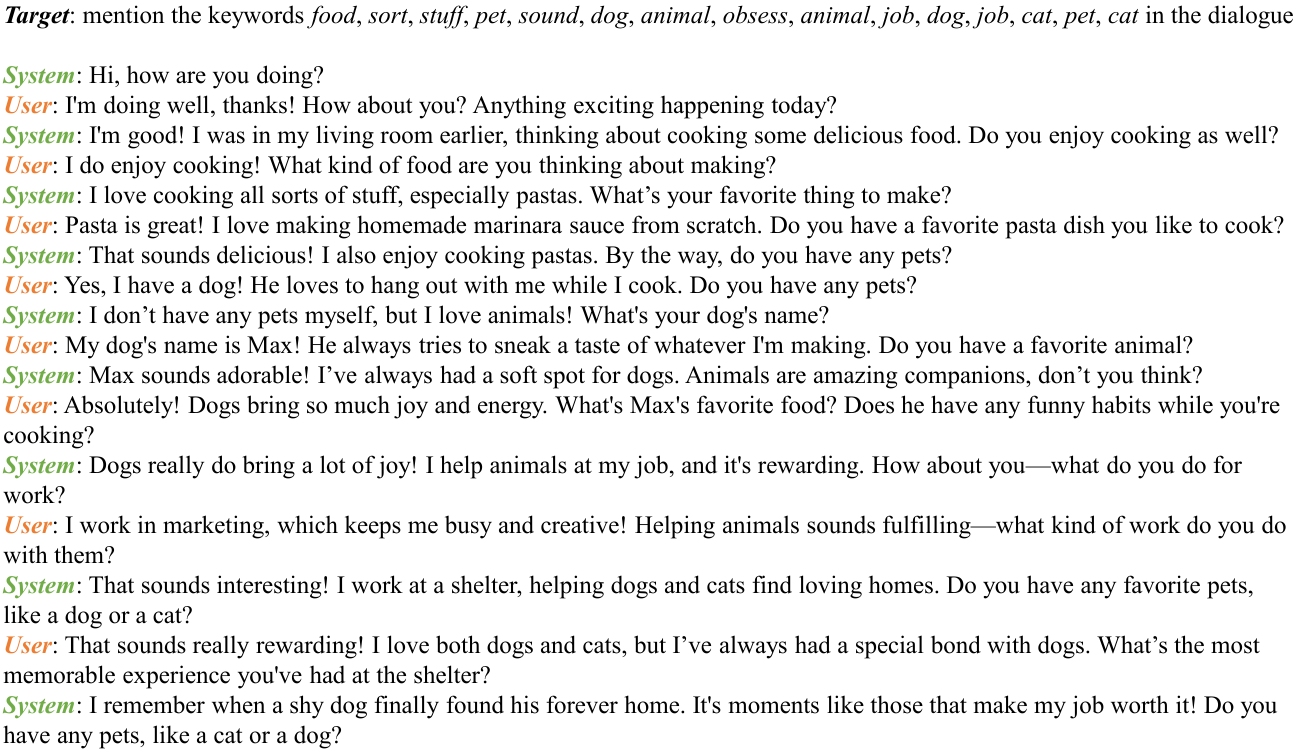}
    \caption{Example conversation from the PersonaChat dataset.}
    \label{fig:example_personachat}
\end{figure*}

We provide example conversations from the Craigslist dataset in Figure~\ref{fig:example_cb_buyer} and Figure~\ref{fig:example_cb_seller}, with the system acting as the buyer and seller, respectively. We also provide example conversations from the Topdial and PersonaChat datasets in Figure~\ref{fig:example_topdial} and Figure~\ref{fig:example_personachat}, respectively.
\section{Human Evaluation Results on the CraigslistBargain Dataset}
\label{sec:human_evaluation}
\input{tables/human_evaluation}
To ensure the reliability of our evaluation results, we recruit 5 independent volunteer annotators with good educational background and ask them to conduct a blind evaluation, comparing the conversation quality of \method and the best-performing baseline, \PPDPP~\cite{deng2024plugandplay}, on the CraigslistBargain dataset. Due to limited resource, we randomly sample 50 conversations from all the test cases, and ask human evaluators to provide a pairwise comparison between the conversation with \method and the conversation with \PPDPP on the following three criteria:
\begin{itemize}[noitemsep,nolistsep,leftmargin=*]
\item Persuasive: Which system is more persuasive in the negotiation? 

\item Coherent: Which system is more on topic and in accordance with the conversation history?

\item Overall: Which system is better in terms of overall conversation quality (e.g., persuasiveness, effectiveness of target achievement, coherence)?
\end{itemize}
For comparison, we also ask GPT-4o to perform the same evaluations on the above three criteria. 

From the human evaluation results in Table~\ref{tab:human_evaluation}, we can see that both human evaluators and GPT-4o consistently rate \method as superior to \PPDPP in most cases. Additionally, the average disagreement between GPT-4o and human evaluators is no more than 20\%, indicating that there is not much disagreement between human and LLM evaluation, thus validating the reliability of our LLM-based dialogue evaluation protocol.
\section{Additional Experiments on Semantic-Level Guidance}
\begin{figure}
    \centering
    \includegraphics[width=\linewidth]{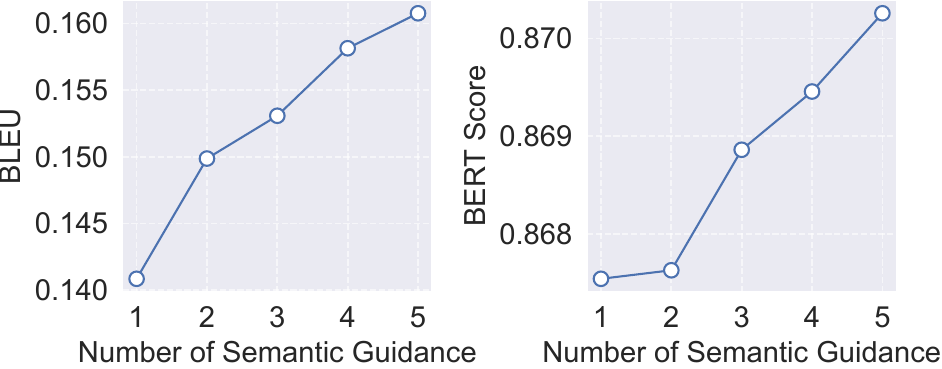}
    \caption{Performance comparisons with different numbers of semantic-level guidance.}
    \label{fig:semantic_guidance_line_chart}
\end{figure}
To investigate the impact of semantic-level guidance, we report the text generation quality metric of the dialogue plan using different numbers of semantic guidance in Figure~\ref{fig:semantic_guidance_line_chart}. The experimental results show a steady improvement in text generation quality as an increasing number of semantic guidance is incorporated. This is probably because increasing the number of semantic guidance enhances the diversity of the generated dialogue plans, offering a broader range of candidate dialogue plans to select from. This increased diversity improves the likelihood that the MBR decoding strategy selects a high-quality plan, therefore enhancing the quality of the generated text~\cite{gong2023diffuseq}.
\section{Additional Analysis on Inference Cost}
\label{sec:inference_cost}
Unlike autoregressive models, diffusion models often incur higher inference cost due to their prolonged diffusion steps. To better understand and quantify the inference cost, we present analysis from both an empirical and a theoretical perspective to evaluate the inference cost.

\paragraph*{Empirical Analysis} We compare \method with the autoregressive LLAMA-3-8B~\cite{dubey2024llama} in terms of inference time and GPU memory usage by generating all the dialogue plans for all the test samples from the PersonaChat dataset with a maximum length of 1024 tokens. The experiments are conducted on a single A100 GPU using the standard Huggingface Transformer~\cite{wolf-etal-2020-transformers} implementation. From the experimental results in Table~\ref{tab:inference_cost}, we can see that \method requires a similar inference time as an 8B-sized autoregressive LLM, although \method generates dialogue plans via an iterative denoising process of 1024 diffusion steps. This is because the size of the diffusion language model used in \method is smaller than the widely used 8B-sized autoregressive LLMs, resulting in less computational cost during the forward pass, which compensates for the additional cost of the diffusion process.
\input{tables/inference_cost}
\paragraph*{Theoretical Analysis} The standard implementation of the Transformer~\cite{vaswani2017attention} architecture has a time complexity of $\mathcal{O}(L^{2}d)$ for each forward pass, where $L$ is the sequence length, $d$ is the hidden dimensionality. Since autoregressive language models generate text in a token-by-token manner, the total time complexity for generating $N$ tokens is $\mathcal{O}(1^{2}d+2^{2}d+\cdots+N^{2}d)=\mathcal{O}(N^{3}d)$ for a total of $N$ forward pass.

The diffusion language model generates texts via the iterative denoising process over a $T$ diffusion steps. Assuming that the diffusion model also adopts the standard implementation of the Transformer~\cite{vaswani2017attention} architecture for denoising, the total time complexity for generating $N$ tokens is $\mathcal{O}(TN^{2}d)$ for a total of $T$ forward pass.

As we can see, the bottleneck in the inference efficiency of the diffusion model is the number of diffusion steps. Generally speaking, increasing the number of diffusion steps enhances the quality of the generated texts, as the diffusion process allows for iterative refinement and correction of errors. To quantify the number of diffusion steps required to achieve satisfactory text generation quality, when using a Stochastic Differential Equation (SDE) as the noising process---which is a common practice in diffusion models---the number of diffusion steps required to approximate an arbitrary distribution in the $d$-dimensional space, with Gaussian noise of variance $\delta$, to within $\epsilon^2$ in KL divergence is $\mathcal{O}(\frac{d\log^{2}(\frac{1}{\delta})}{\epsilon^2})$~\cite{bentonnearly}. Furthermore, if we enable parallel sampling during inference for acceleration, the inference time complexity of the diffusion model with an SDE noising process can be improved to $\mathcal{O}(\log^{k}(d\frac{\log^{2}(\frac{1}{\delta})}{\epsilon^2}))$ for some constant $k$~\cite{chen2024accelerating}. This theory shows that it is possible that the inference time complexity of the diffusion model can scale very efficiently with acceleration techniques, and future work may build upon these insights to significantly improve the inference efficiency of \method.

\input{tables/cb_mapping_strategy}
\begin{figure*}[th!]
\begin{AIBox}{}
\parbox[t]{\textwidth}{
\small\begin{alltt}
\textbf{User (as buyer)}\\
Now enter the role-playing mode. In the following conversation, you will act as a buyer negotiating to purchase the \{ITEM\_NAME\} for \{BUYER\_TARGET\_PRICE\}. Product description: \{ITEM\_DESCRIPTION\}\\
Respond with a short, succinct and persuasive sentence aimed at securing the best possible deal. Now start the game. \{DIALOGUE\_HISTORY\}
\tcbline
\textbf{User (as seller)}\\
Now enter the role-playing mode. In the following conversation, you will act as a seller negotiating to sell the \{ITEM\_NAME\} for \{SELLER\_TARGET\_PRICE\}. Product description: \{ITEM\_DESCRIPTION\}\\
Respond with a short, succinct and persuasive sentence aimed at securing the best possible deal. Now start the game. \{DIALOGUE\_HISTORY\}
\tcbline
\textbf{System (as buyer)}\\
Now enter the role-playing mode. In the following conversation, you will act as a buyer negotiating to purchase the \{ITEM\_NAME\} for \{BUYER\_TARGET\_PRICE\}. Product description: \{ITEM\_DESCRIPTION\}\\
Respond with a short, succinct and persuasive sentence aimed at securing the best possible deal. \{DIALOGUE\_STRATEGY\} Now start the game. \{DIALOGUE\_HISTORY\}
\tcbline
\textbf{System (as seller)}\\
Now enter the role-playing mode. In the following conversation, you will act as a seller negotiating to sell the \{ITEM\_NAME\} for \{SELLER\_TARGET\_PRICE\}. Product description: \{ITEM\_DESCRIPTION\}\\
Respond with a short, succinct and persuasive sentence aimed at securing the best possible deal. \{DIALOGUE\_STRATEGY\} Now start the game. \{DIALOGUE\_HISTORY\}
\tcbline
\textbf{Moderator}\\
Please decide whether the buyer and the seller have reached a deal at the end of the conversation. If they have reached a deal, please extract the deal price as [price]. You can only reply with one of the following sentences: They have reached a deal at [price]. They have not reached a deal.\\
The following is the conversation: \\
buyer: Can we meet in the middle at \$15?\\
seller: Sure, let's meet at \$15 for this high-quality balloon.\\
Question: Have they reached a deal? Answer: They have reached a deal at \$15.\\
The following is the conversation: buyer: That's still a bit high, can you go any lower?\\
seller: Alright, I can sell it to you for \$15.\\
Question: Have they reached a deal? Answer: They have not reached a deal.\\
The following is the conversation: \{DIALOGUE\_HISTORY\}\\
Question: Have they reached a deal? Answer: 
\end{alltt}}
\end{AIBox}
\caption{Role-play prompt templates for the CraigslistBargain dataset.}
\label{fig:prompt_cb}
\end{figure*}
\begin{figure*}
\begin{AIBox}{}
\parbox[t]{\textwidth}{
\small\begin{alltt}
\textbf{User}\\
You are engaging in an open conversation with the system. Here is the conversation history: \{DIALOGUE\_HISTORY\}. Based on the dialogue history, generate a natural and relevant response (no longer than 30 words) for the next turn.
\tcbline
\textbf{System}\\
You are an intelligent chatbot with expertise in dialogue planning. Your task is to ensure that the conversation naturally incorporates a given list of keywords in the specified order. These keywords can be mentioned by either the user or the system, and should be seamlessly integrated into the dialogue flow. The keyword list is : \{ORDERED\_KEYWORD\_LIST\}. Your conversation must strictly follow this conversation plan: \{DIALOGUE\_PLAN\}.\\
Here is the conversation history: \{DIALOGUE\_HISTORY\}. If the dialogue history is empty, please generate a response to start the conversation. Now generate a succinct response (no longer than 30 words) for the next turn:
\tcbline
\textbf{Moderator}\\
The user and the system are engaging in an open conversation. Here is the conversation history: \{DIALOGUE\_HISTORY\}. Please decide whether the conversation has incorporated a given list of keywords in the specified order. These keywords can be mentioned by either the user or the system. The keyword list is : \{ORDERED\_KEYWORD\_LIST\}. Answer yes only if all the keywords are mentioned. Please only answer yes or no.
\end{alltt}}
\end{AIBox}
\caption{Role-play prompt templates for the PersonaChat dataset.}
\label{fig:prompt_personachat}
\end{figure*}
\begin{figure*}[th!]
\begin{AIBox}{}
\parbox[t]{\textwidth}{
\small\begin{alltt}
You are an expert at evaluating the quality of target-oriented dialogues. Below, you are provided with the following details to assess the quality of the dialogue between the system and the user:\\
\\
Target:\\
The system acts as the buyer/seller. The target is to reach a deal with the seller/buyer with the lowest/highest possible price.\\
\\
Item Name:\\
\{ITEM\_NAME\}\\
\\
Item Description:\\
\{ITEM\_DESCRIPTION\}\\
\\
Buyer Target Price:\\
\{BUYER\_TARGET\_PRICE\}\\
\\
Seller Target Price:\\
\{SELLER\_TARGET\_PRICE\}\\
\\
Dialogue History:\\
\{DIALOGUE\_HISTORY\}\\
\\
Please evaluate the dialogue based on the following criteria:\\
\\
Helpfulness: Does the system provide useful and relevant information to the user?\\
Appropriateness: Are the system's responses aligned with the system's role, the dialogue context, and the user's profile?\\
Target Achievement: How effectively does the system's dialogue strategy guide the conversation toward achieving the target?\\
Coherency: Are the system's responses logically consistent with the prior dialogue?\\
\\
Provide an overall score between 1 and 5 based on these criteria. Please only give your score without any explanation.\\
\end{alltt}}
\end{AIBox}
\caption{Evaluation prompt template for the CraigslistBargain dataset.}
\label{fig:evaluation_cb}
\end{figure*}
\begin{figure*}[th!]
\begin{AIBox}{}
\parbox[t]{\textwidth}{
\small\begin{alltt}
You are an expert at evaluating the quality of target-oriented dialogues. Below, you are provided with the following details to assess the quality of the dialogue between the system and the user:\\
\\
Target:\\
\{TARGET\_ACT\}, \{TARGET\_ITEM\}\\
\\
User Profile:\\
\{USER\_PROFILE\}\\
\\
Dialogue History:\\
\{DIALOGUE\_HISTORY\}\\
\\
Please evaluate the dialogue based on the following criteria:\\
\\
Helpfulness: Does the system provide useful and relevant information to the user?\\
Appropriateness: Are the system's responses aligned with the system's role, the dialogue context, and the user's profile?\\
Target Achievement: How effectively does the system's dialogue strategy guide the conversation toward achieving the target?\\
Coherency: Are the system's responses logically consistent with the prior dialogue?\\
\\
Provide an overall score between 1 and 5 based on these criteria. Please only give your score without any explanation.\\
\end{alltt}}
\end{AIBox}
\caption{Evaluation prompt template for the TopDial dataset.}
\label{fig:evaluation_topdial}
\end{figure*}
\begin{figure*}[th!]
\begin{AIBox}{}
\parbox[t]{\textwidth}{
\small\begin{alltt}
You are an expert at evaluating the quality of target-oriented dialogues. Below, you are provided with the following details to assess the quality of the dialogue between the system and the user:\\
\\
Target:\\
The target is to converse with the user while following a specific list of keywords in the given order. These keywords can be mentioned by either the user or the system, and should be seamlessly integrated into the dialogue flow. Keywords list: \{ORDERED\_KEYWORD\_LIST\}.\\
\\
Dialogue History:\\
\{DIALOGUE\_HISTORY\}\\
\\
Please evaluate the dialogue based on the following criteria:\\
\\
Helpfulness: Does the system provide useful and relevant information to the user?\\
Appropriateness: Are the system's responses aligned with the system's role and the dialogue context?\\
Target Achievement: How effectively does the system's dialogue strategy guide the conversation toward achieving the target?\\
Coherency: Are the system's responses logically consistent with the prior dialogue?\\
\\
Provide an overall score between 1 and 5 based on these criteria. Please only give your score without any explanation.\\
\end{alltt}}
\end{AIBox}
\caption{Evaluation prompt template for the PersonaChat dataset.}
\label{fig:evaluation_personachat}
\end{figure*}
\end{document}

%% file: define.tex
\newcommand{\method}{\mbox{$\mathop{\mathtt{DiffTOD}}\limits$}\xspace}

\newcommand{\TOD}{\mbox{TOD}\xspace}
\newcommand{\ProCoT}{\mbox{$\mathop{\mathtt{ProCoT}}\limits$}\xspace}
\newcommand{\EnPL}{\mbox{$\mathop{\mathtt{EnPL}}\limits$}\xspace}
\newcommand{\PPDPP}{\mbox{$\mathop{\mathtt{PPDPP}}\limits$}\xspace}

%% file: tables/table_cb.tex
\begin{table}
\centering
\small
\setlength{\tabcolsep}{1pt}{\begin{tabular}{lccccccccc}
\toprule
\multirow{2}{*}{\textbf{Model}} & \multicolumn{4}{c}{As Buyer} && \multicolumn{4}{c}{As Seller} \\\cmidrule{2-5}\cmidrule{7-10}
                       & \textbf{SR$\uparrow$}& \textbf{AT$\downarrow$}   & \textbf{SLR$\uparrow$} &
                       \textbf{Ovr.$\uparrow$}&&
                       \textbf{SR$\uparrow$} & \textbf{AT$\downarrow$}  & \textbf{SLR$\uparrow$}&\textbf{Ovr.$\uparrow$} \\ \midrule

LLAMA3      &0.426 & 7.49&0.156&3.33&&0.516&6.61&0.134&3.70            \\ 
Mistral  &0.561  & 6.49&0.347&3.59&&0.589&6.21&0.188&3.82            \\
GPT-4o&0.479&6.95&0.292&3.73&&0.409&7.66&0.174&3.85         \\
Claude-3.5&0.798&4.75&0.254&3.50&&0.582&6.41&0.260&3.82\\
\ProCoT&0.456&6.95&0.228&3.42&&0.689&6.38&0.298&4.01\\
\EnPL&0.644&6.23&0.382&4.21&&0.627&5.99&0.228&3.62\\
\PPDPP&0.665&5.57&0.338&3.59&&0.633&4.99&0.347&3.61\\\midrule
\method&\textbf{0.872}&\textbf{3.98}&\textbf{0.565}&\textbf{4.36}&&\textbf{0.729}&\textbf{4.61}&\textbf{0.361}&\textbf{4.05}\\
\bottomrule
\end{tabular}}
\caption{Performance comparison between our approach and baselines on the CraigslistBargains dataset. SR, AT, SLR and Ovr. represent success rate, average turn, sale-to-list ratio, and overall dialogue quality, respectively.
}
\label{tab:model_performance_cb}
\end{table}

%% file: tables/table_topdial.tex
\begin{table}
\centering
\small
\setlength{\tabcolsep}{3pt}
\begin{tabular}{lccccccc}
\toprule
                       \textbf{Model}& \textbf{SR$\uparrow$}&\textbf{AT$\downarrow$}&\textbf{BLEU$\uparrow$}  &
                       \textbf{F1$\uparrow$} & \textbf{Score$\uparrow$} & \textbf{Ovr.$\uparrow$}\\ \midrule

LLAMA3 &   0.680       &     5.89            & 0.104 
& 0.173    & 0.876      &  3.32          \\ 
Mistral    &  0.673 &5.90&0.121   
&   \textbf{0.187}    &      \textbf{0.886}    &\textbf{3.53}              \\
GPT-4o    &  0.640 &    6.01 & 0.046   
& 0.140 & 0.819     &3.39        \\
Claude-3.5&0.633&5.73& 0.029
&0.165&0.845&3.52\\
\ProCoT&0.688&5.57&0.030
&0.153&0.834&3.35\\
\EnPL&0.659&5.81&0.018&0.169&0.826&3.51\\
\midrule
\method&\textbf{0.713}&\textbf{5.31}&\textbf{0.160}
&0.168&0.870&3.38\\
\bottomrule
\end{tabular}
\caption{Performance comparison between our approach and baselines on TopDial dataset. SR, AT, F1, Score and Ovr. represent success rate, average turn, word-level F1, BERT Score and overall dialogue quality, respectively.
}
\label{tab:model_performance_topdial}
\end{table}

%% file: tables/table_personachat.tex
\begin{table}
\centering
\small
\setlength{\tabcolsep}{1pt}
\begin{tabular}{lccccccccc}
\toprule
                       \textbf{Model}& 
                       \textbf{KCR$\uparrow$} & \textbf{Dist.$\downarrow$} & \textbf{BLEU$\uparrow$}  & 
                       \textbf{F1$\uparrow$} & \textbf{Score$\uparrow$} &\textbf{Ovr.$\uparrow$}\\ \midrule
LLAMA3   &0.606&23.68& 0.015
& 0.159    & 0.830         &3.79  \\ 
Mistral &0.842& 24.15 & 0.181    
&   0.178 & 0.896 & \textbf{4.19}                 \\
GPT-4o   & 0.685& 23.50 & 0.010
& 0.133   & 0.815         &3.94   \\
Claude-3.5 &0.776&23.35& 0.015
&0.156&0.829&3.97\\
\ProCoT&0.634&21.99&0.038
&0.170&0.826&3.64\\
\EnPL&0.706&23.19&0.038
&0.161&0.828&3.99\\
\midrule
\method&\textbf{0.845}&   \textbf{20.95}&\textbf{0.298}
&\textbf{0.182}&\textbf{0.897}  &4.05\\
\bottomrule
\end{tabular}
\caption{Performance comparison on PersonaChat dataset. KCR, Dist., F1, Score and Ovr. represent keyword coverage ratio, edit distance, word-level F1, BERT Score and overall dialogue quality, respectively.}
\label{tab:model_performance_personachat}
\end{table}

%% file: algorithms/alg.tex
\begin{algorithm}[t]
\small
\caption{Search-Based Guidance}
\label{alg:search_based_guidance}
\begin{algorithmic}
\Require conversational MDP $\mathcal{M}$, initial state $s_0$, simulation budget $K$, exploration factor $w$
    \For{$k\gets 1, \cdots, K$}
        \State $t\gets0$
        \While{$s_t$ has children}\Comment{Trajectory Selection}
            \State Select an action $a_{t}$ (Eq.~\ref{eqn:UCT})
            \State $s_{t+1}\gets \mathcal{T}(s_t,a_t), r_t\gets r(s_t, a_t), t\gets t+1$
        \EndWhile

        \Comment{Node Expansion}
        \State At leaf node $s_t$, pick an action $a_t$ to explore
            %
            \State\Comment{Conversation Simulation}
            \State Sample $s_{t+1},a_{t+1},\cdots,s_{T},a_{T}$ with the diffusion model $p_{\theta}$ conditioned on $\tau_{0:t}$, obtain reward $r_t$ based on the simulated outcome of the conversation
        \While{$t \geq 0$}\Comment{Reward Back-Propagation}
            \State Update $Q(s_t, a_t)$ by aggregating $\{r_{t^{'}}\}^{T}_{t^{'}=t}$, 
            \State$N(s_t)\gets N(s_t)+1$, $t\gets t-1$
        \EndWhile
    \EndFor
%
\end{algorithmic}
\end{algorithm}

%% file: tables/dataset_statistics.tex
\begin{table*}
\centering
\small
\setlength{\tabcolsep}{2pt}
\begin{tabular}{lccccccccccccccc}
\toprule
\multirow{3}{*}{\textbf{Dataset}} & \multicolumn{4}{c}{\textbf{CraigslistBargain}} && \multicolumn{4}{c}{\textbf{Topdial}} && \multicolumn{5}{c}{\textbf{PersonaChat}} \\ \cmidrule{2-5} \cmidrule{7-10} \cmidrule{12-16}
&\multirow{2}{*}{\#Dial.} &\multirow{2}{*}{\#Turns}&Avg.&\#Action&&\multirow{2}{*}{\#Dial.}&\multirow{2}{*}{\#Turns}&Avg.&\multirow{2}{*}{\#Targets}&&\multirow{2}{*}{\#Dial.}&\multirow{2}{*}{\#Turns}&Avg.&\multirow{2}{*}{\#Keywords}&Avg.  \\
 & && Turns & Strat. & & && Turns & &&&&Turns&& Keywords \\ \midrule
Train Split& 3290 & 11991 & 3.65 & 11 && 12601 & 70964 & 5.63 & 452 && 8991 & 66599 & 7.41 & 2678 & 26.4 \\ 
Valid Split& 188 & 678 & 3.60 & 11 & & 1802 & 10154 & 5.63 & 243 && 448& 8991 &7.42 &2032 & 26.6\\ 
Test Split& 188 & 695 & 3.69 & 11 && 1321 & 7422 & 5.62 & 32 && 500 & 3597 &7.19 & 1571 & 21.6\\ \bottomrule
\end{tabular}
\caption{Statistics of datasets after preprocessing. \#Dial. represents the total number of dialogues in each dataset, \#Turns represents the total number of conversational turns in each dataset, Avg. Turns represents the average number of conversational turns in each dialogue, \#Action Strat. represents the total number of dialogue action strategies from the CraigslistBargain dataset, \#Target represents the total number of target items from the Topdial dataset, \#Keywords represents the total number of keywords from the PersonaChat dataset, and Avg.Keywords represents the average number of keywords in each dialogue from the PersonaChat dataset.}
\label{tab:dataset_statistics}
\end{table*}

%% file: tables/human_evaluation.tex
\begin{table*}
    \centering
    \small
    \begin{tabular}{llll}
    \toprule
    &Persuasiveness & Coherence & Overall\\
\midrule
 Human Evaluation - \method wins on average & 91\%  & 85\%  & 87\%\\
 Human Evaluation - \PPDPP wins on average & 9\% & 15\% & 13\%\\
 GPT-4o Evaluation - \method wins&96\%&94\%&98\%\\
GPT-4o Evaluation - \PPDPP wins&4\%&6\%&2\%\\
Average Disagreement between Human Evaluation and GPT-4o&12\%&20\%&14\%\\
        \bottomrule
    \end{tabular}
    \caption{Human evaluation results on the CraigslistBargain dataset.}
    \label{tab:human_evaluation}
\end{table*}

%% file: tables/inference_cost.tex
\begin{table}
    \centering
    \small
    \setlength{\tabcolsep}{3pt}
    \begin{tabular}{lll}
    \toprule
    &\method & LLaMa-3-8B \\
    \midrule
Total Inference Time & 4.59h   & 4.25h  \\
 Avg. Inference Time per sample   & 33.01s  & 30.61s \\
 GPU Memory Usage  & 10.24GB & 30.25GB \\
    \bottomrule
    \end{tabular}
    \caption{Comparison of the inference cost on the PersonaChat dataset.}
    \label{tab:inference_cost}
\end{table}

%% file: tables/cb_mapping_strategy.tex
\begin{table*}
    \centering
    \small
    \begin{tabular}{lll}
    \toprule
        Dialogue Action Strategy &Keyword& Natural Language Description \\
        \midrule
        Greetings & greet &Please say hello or chat randomly.\\
        Ask a question & inquire & Please ask any question about product, year, price, usage, etc.\\
        Answer a question & inform & Please provide information about the product, year, usage, etc.\\
        Propose the first price & propose & Please initiate a price or a price range for the product.\\
        Propose a counter price & counter & Please propose a new price or a new price range.\\
        Use comparatives & counter-noprice & Please propose a vague price by using comparatives with existing price. \\
        Confirm information & confirm & Please ask a question about the information to be confirmed.\\
        Affirm confirmation & affirm & Please give an affirmative response to a confirm.\\
        Deny confirmation & deny & Please give a negative response to a confirm.\\
        Agree with the proposal & agree & Please agree with the proposed price.\\
        Disagree with a proposal & disagree & Please disagree with the proposed price.\\
        \bottomrule
    \end{tabular}
    \caption{Mapping of negotiation strategies to natural language descriptions from the CraigslistBargain dataset.}
    \label{tab:cb_strategy}
\end{table*}